\documentclass[journal]{IEEEtran/IEEEtran}

\ifCLASSOPTIONcompsoc
  \usepackage[caption=false,font=normalsize,labelfont=sf,textfont=sf]{subfig}
\else
 \usepackage[caption=false,font=footnotesize]{subfig}
\fi

\usepackage{amssymb} %spconf
\usepackage{booktabs, multicol, multirow, bigstrut, array}
\newcolumntype{R}[1]{>{\raggedleft\let\newline\\\arraybackslash\hspace{0pt}}m{#1}}
\usepackage{hhline}
\usepackage{cite}

\usepackage[dvips]{graphicx}
\DeclareGraphicsExtensions{.eps,.bmp}
\usepackage{pgf}
\usepackage{tikz}
\usepackage{pstricks} 
\usepackage[cmex10]{amsmath}
\usepackage{fixmath}
\usepackage{fixltx2e}
\usepackage{hyperref}

\usepackage[flushleft]{threeparttable}

\DeclareMathOperator{\E}{\mathbb{E}}
\usepackage{float}
\begin{document}
\title{Unpaired Speech Enhancement by Acoustic and Adversarial Supervision for Speech Recognition}
%\title{Robustness to Spectral Variations of Speech by\newline
%an Intermap Pooling Layer in a Deep CNN }


\author{Geonmin Kim, Hwaran Lee, Bo-Kyeong Kim, Sang-Hoon Oh, and Soo-Young Lee
         
\thanks{This work was supported by the Industrial Strategic Technology Development Program (10076757, Free-Running Embedded Speech Recognition Technology for Natural Language Dialogue with Robots) funded by the Ministry of Trade, Industry and Energy (MOTIE, Korea)}%
\thanks{
G. Kim, H. Lee and B.-K. Kim are with the Department of Electrical Engineering, KAIST, Daejeon 305-701, Korea. (e-mail:
gmkim90@gmail.com, hwaran.lee@kaist.ac.kr, bokyeong1015@gmail.com)}
\thanks{
S.-H Oh is with the Division of Information and Communication Convergence
Engineering, Mokwon University, Daejeon, 302-318, Korea. (e-mail:
shoh@mokwon.ac.kr)}
\thanks{
S.-Y. Lee is with the KAIST Institute for Artificial Intelligence, Daejeon, 305-701, Korea. (e-mail: sy-lee@kaist.ac.kr)}
}

%\markboth{Journal of \LaTeX\ Class Files,~Vol.~13, No.~9, September~2014}%
%{Shell \MakeLowercase{\textit{et al.}}: Bare Demo of IEEEtran.cls for Journals}



\maketitle


\begin{abstract}
% Ver1
%Recent speech enhancement methods relying on clean speech target has several limitations: simulated speech requiring extensive environment information, simulated speech may not include a human behavioral pattern, and mismatch between speech intelligibility and speech recognition. We propose Acoustic and Discriminator Supervision (ADS) as a learning algorithm of speech enhancement which directly improve speech recognition performance with having less artifact. ADS is tested on simulated (Librispeech + DEMAND) and real (CHiME-4) noisy dataset. It outperforms other methods relying on clean speech target and proposed system achieve 24.9\% WER on CHiME-4 real test set.

% Ver2
Many speech enhancement methods try to learn the relationship between noisy and clean speech, obtained using an acoustic room simulator. We point out several limitations of enhancement methods relying on clean speech targets; the goal of this work is proposing an alternative learning algorithm, called acoustic and adversarial supervision (AAS). AAS makes the enhanced output both maximizing the likelihood of transcription on the pre-trained acoustic model and having general characteristics of clean speech, which improve generalization on unseen noisy speeches. We employ the connectionist temporal classification and the unpaired conditional boundary equilibrium generative adversarial network as the loss function of AAS. AAS is tested on two datasets including additive noise without and with reverberation, Librispeech + DEMAND and CHiME-4. By visualizing the enhanced speech with different loss combinations, we demonstrate the role of each supervision. AAS achieves a lower word error rate than other state-of-the-art methods using the clean speech target in both datasets.
\end{abstract}


% \ifCLASSOPTIONpee rreview
% \fi
\IEEEpeerreviewmaketitle

\section{Introduction}
\label{sec:intro}
\IEEEPARstart{T}{echniques} for single-channel speech enhancement range from conventional signal processing methods such as minimum mean square error \cite{ephraim84}, wiener filter \cite{pascal96}, and subspace algorithm \cite{ephraim95} to expressive deep neural network \cite{pascual17, rethage18, omologo17}. Most of the latter approaches are based on supervised learning, which requires clean speech paired with the noisy mixture to learn the relationship between them. Since such pairs are generally unknown, they need to be generated artificially from clean speech, assuming that they will match the target noisy environment. However, speech enhancement methods relying on clean speech targets have several limitations.

Firstly, the acoustic room simulator requires extensive environment information (i.e., room size distribution, reverberation time, source to microphone distance, and noise type) \cite{kim17} to convolve the room impulse response and add noise to the clean speech. This information can be estimated from noisy speech; however, this itself is a challenging problem \cite{lafay17, baba18}.

Secondly, the acoustic model trained on simulated data is often not generalized well in a real environment \cite{vincent16}. This is because simulation may not fully cover the real environment or represent characteristics other than additive noise and reverberation (e.g., Lombard effect \cite{junqua99}). 

Thirdly, when enhancement is used as a preprocessing stage for speech recognition, enhancement towards clean speech may not be the optimal approach. Speech recognition requires the phonetic characteristics in the enhanced speech to be preserved while suppressing other non-verbal details. However, yielding enhanced outputs that resemble clean speech is different from this direction.

To avoid the use of clean speech targets, we propose an alternative learning algorithm: acoustic and adversarial supervision (AAS). Acoustic supervision teaches an enhancement model to yield outputs that are recognized correctly by the pre-trained acoustic model. Adversarial supervision trains the enhancement model to yield outputs that have the general characteristics of clean speech. AAS \footnote{Code and the supplmentary results are available at \url{https://github.com/gmkim90/AAS_enhancement}.} is compared with other state-of-the-art methods using clean speech target in synthetic and real noisy datasets.
The remainder of this paper describes the review on conditional generative adversarial networks, the proposed AAS algorithm, experimental setting, and results.
\section{Conditional Generative Adversarial Network for Speech Enhancement}
\label{sec:cGAN}
Speech enhancement is related to domain transfer problems (e.g., image-to-image \cite{liu17} and voice conversion\cite{hsu17}) where the source and target domains are the noisy and clean recording environments, respectively. The representative work is the frequency speech enhancement generative adversarial network (FSEGAN, \cite{donahue17}) which employs two losses: the distance from the clean to enhanced speech and the loss function for the conditional generative adversarial network (cGAN, \cite{mirza15}). Given a source domain ($\textbf{x}_{s}$) and a target domain ($\textbf{x}_{t}$) data, cGAN optimizes the min-max game between a generator ($G$) and a discriminator ($D$) with the value function ($V$) given by
\begin{equation}
\begin{split}
      \min_{G}\max_{D} V_{cGAN}(G, D) = \E_{(\textbf{x}_{s}, \textbf{x}_{t}) \sim p(\textbf{x}_{s}, \textbf{x}_{t})}[\log D(\textbf{x}_s, \textbf{x}_{t})] \\ +  \E_{\textbf{x}_{s} \sim p(\textbf{x}_{s}), \textbf{z} \sim N(0, I)}[\log (1-D(\textbf{x}_{s}, G(\textbf{x}_{s}, \textbf{z})))]. 
\end{split}
\end{equation}
Here, $G$ is trained to deceive $D$, which judges whether a given pair of cross-domain samples come from the real data $(\textbf{x}_{s},\textbf{x}_{t})$ or are generated from the source domain and random noise $\textbf{z}$ $(\textbf{x}_{s}, G(\textbf{x}_{s}, \textbf{z}))$. Two losses of FSEGAN require the paired clean and noisy speeches, not available in the real environment.

Usually, domain transfer problems require unsupervised learning because the paired data between different domains are expensive to be obtained. Therefore, many domain transfer models based on cGAN \cite{liu17,zhu17,kim17_2} remove the dependency of the paired source in a discriminator and use the unpaired cGAN (upcGAN) whose value function ($V$) is
\begin{equation}
\begin{split}
      \min_{G}\max_{D}V_{upcGAN}(G, D) = \\ \E_{\textbf{x}_{t} \sim p(\textbf{x}_{t})}[\log D(\textbf{x}_{t})] +\E_{\textbf{x}_{s} \sim p(\textbf{x}_{s})}[\log (1-D(G(\textbf{x}_{s})))],
\end{split}
\end{equation}
where $\textbf{z}$ is often omitted to learn deterministic generator.

However, upcGAN can lead the transferred sample $G(\textbf{x}_{s})$ merely having the general characteristics of the target domain, since the discriminator judges the transferred sample without seeing the paired source domain sample. This problem can be alleviated by imposing additional regularization \cite{kwak16} on a generated sample, such as cycle-consistency loss \cite{zhu17, kim17_2}. However, this loss is not applicable for speech enhancement because the original noisy speech cannot be reconstructed from enhanced speech since there are infinite possible noises to mix. Instead, we encourage the enhanced sample to be recognized correctly by an acoustic model as an alternative regularization.

\section{Acoustic and Adversarial Supervision}
\label{sec:MTL}
\begin{figure}[t]
 \centering
  \includegraphics[width=8cm] {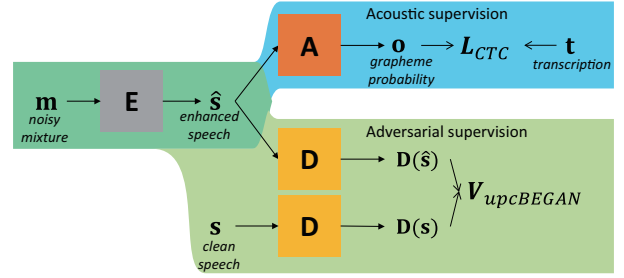}
   \caption{The proposed acoustic and adversarial supervision (AAS). The enhancement model (E) is trained with two loss functions: the acoustic supervision ($L_{CTC}$) computed using the acoustic model (A) and the adversarial supervision ($V_{upcBEGAN}$) computed using the discriminator (D).}
 \label{proposed_method}
\end{figure}

 We propose acoustic and adversarial supervision (AAS) for a speech-enhancement learning algorithm, as shown in Fig. \ref{proposed_method}. The proposed method consists of three models: Enhancement (E), Acoustic (A), and Discriminator (D). For the following description, \textbf{m}, \textbf{s}, $\hat{\textbf{s}}$, \textbf{o}, and \textbf{t} are the noisy mixture, (unpaired) clean speech, enhanced speech, grapheme probability, and transcription, respectively. We assume $\textbf{s}$ and pairs of  $(\textbf{m}, \textbf{t})$ are available for the training data.

\subsection{Acoustic Supervision}
\label{ssec:SR task}
 Acoustic supervision trains the enhancement model to maximize the likelihood of transcription of the noisy sample. The pre-trained acoustic model (AM) provides the enhancement model with top-down information of the phonetic features essential for correct recognition. This is motivated from top-down attention mechanism of humans, applied for noise-robust speech recognition \cite{lee07}, and N-best rescoring \cite{kim18}. Although this supervision does not require a specific type of AM, we employ a neural network with connectionist temporal classification (CTC, \cite{graves06}). The CTC is used to label a sequence without requiring explicit alignment between the input and label sequences. Moreover, grapheme is used as the output unit of the neural network, so that AM does not require a lexicon, which allows generating out-of-vocabulary words during inference. The CTC loss function is given by
\begin{equation}
      L_{CTC}(E) =  -\E_{(\textbf{m}, \textbf{t}) \sim p(\textbf{m}, \textbf{t})} [\text{log}  p(\textbf{t}|E(\textbf{m}))],
\end{equation}
\begin{equation}
		p(\textbf{t}|E(\textbf{m})) =  \sum_{\textbf{$\pi$} \in Align(E(\textbf{m}), \mathbf{\tilde{t}})}\prod_{f} o_{f}^{\pi},
\end{equation}
where $\mathbf{\tilde{t}}$ is a sequence with CTC-blank added between every pair of graphemes in $\textbf{t}$, the beginning, and the end. 
The likelihood of $\textbf{t}$ given $E(\textbf{m})$ is defined as sum of single path likelihoods across all possible alignments ($Align(E(\textbf{m}), \mathbf{\tilde{t}})$).

\subsection{Adversarial Supervision}
\label{ssec: DT task}
Adversarial supervision encourages the enhanced speech to have the characteristics of clean speech. We employ upcGAN by replacing $G$ with $E$. The training convergence of upcGAN is improved further by leveraging the techniques of boundary equilibrium GAN (BEGAN, \cite{berthelot17}).

Firstly, the discriminator ($D$) auto-encodes the inputs ($l_{D}(\textbf{x}) = |\textbf{x}-D(\textbf{x})|$) instead of using binary logistic prediction to enhance training efficiency by providing diverse directions of the gradients within the minibatch \cite{zhao17}. Secondly, to balance the power of the discriminator ($D$) and the enhancement ($E$) model, the importance of loss on the clean sample ($\E_{\textbf{s}\sim p(\textbf{s})}[l_{D}(\textbf{s})]$) is controlled by the proportional control theory \cite{berthelot17} given by formula (6). This control helps to maintain the ratio of loss between clean and enhanced data as the pre-defined constant ($\gamma \in [0,1]$): $\E_{\textbf{m}\sim p(\textbf{m})}[l_{D}(E(\textbf{m}))]/\E_{\textbf{s}\sim p(\textbf{s})}[l_{D}(\textbf{s})] = \gamma$. The final value function for $D$ and $E$ is given by
\begin{equation}
\begin{split}
      \min_{E}\max_{D}V_{upcBEGAN}(E, D) = \\ \E_{\textbf{m}\sim p(\textbf{m})}[l_{D}(E(\textbf{m}))]-1/(k_{t}+\epsilon)\E_{\textbf{s}\sim p(\textbf{s})}[l_{D}(\textbf{s})],
\end{split}
\end{equation}
\begin{equation}
      k_{t+1} = k_{t} + \lambda(\gamma \E_{\textbf{s}\sim p(\textbf{s})}[l_{D}(\textbf{s})]-\E_{\textbf{m}\sim p(\textbf{m})}[l_{D}(E(\textbf{m}))]), 
\end{equation}
where $k_{t} \in [0,1], k_{0} = 0, \epsilon=10^{-8}$.

\subsection{Multi-task Learning}
\label{ssec: MTL}
An enhancement model trained using acoustic supervision directly increases the likelihood of transcription on the AM. However, such a model is not unique and depends on the initialization of model parameters and training data. Due to the non-uniqueness, the enhanced output is not guaranteed to converge towards natural speech and often includes artifacts. Moreover, the optimal parameters differ depending on training data, which may not generalize well on an unseen data.

To constrain the solution, we employ the adversarial supervision as an auxiliary task. The adversarial supervision regularizes the enhanced speech having less artifacts, leading to the improved generalization on an unseen data.

Both losses are combined with weight $w_{AC}$ and $w_{AD}$ as
\begin{equation}
      \min_{E}\max_{D} w_{AC}L_{CTC}(E) + w_{AD}V_{upcBEGAN}(E,D).
\end{equation}
\raggedbottom
\section{Experimental setting}
\label{sec: Experimental setting}
\subsection{Common setting}
\label{ssec: setting}
In all experiments, all the parameters of the neural network are randomly initialized with the distribution $N(0,0.1^2)$. Adam optimizer \cite{kingma15} with learning rate $10^{-5}$ and minibatch size 30 is used for training the model. The performance on the test data is reported when the word error rate (WER) on the validation data is the minimum out of 100 epochs. We use $\gamma = 0.5, \lambda = 0.001$ for optimizing the $V_{upcBEGAN}$. 
 
For the language model (LM), 4-gram trained with the Librispeech text corpus is used.\footnote{The resources are available in \url{http://www.openslr.org/11/}.} 100-best hypotheses, obtained by beam search on AM, are rescored by combining AM score and length normalized word-level LM score \cite{amodei16} given by
\begin{equation}
S = \log p(\textbf{y}|\textbf{x}) + \alpha \log (p(\textbf{y})/|\textbf{y}|^\beta).
\end{equation}

We use 40-dimensional log-mel filterbank (LMFB) feature as the feature for enhancement and recognition. We employ 30 symbols (26 alphabets, underscore, apostrophe, whitespace, and CTC-blank) to represent the AM output.

The AM is trained with the Librispeech corpus \cite{panayotov15}, which provides 960 h of read speech collected from 2,338 speakers as the training data. The AM combined with LM achieves a WER of 5.7\% on the test-clean of Librispeech, which is competitive with DNN-HMM (5.3\%, \cite{panayotov15}). This AM is applied for both the noisy domain datasets described in Section IV-B.

\subsection{Noisy dataset}
\label{ssec: DB}
Librispeech + DEMAND \cite{thiemann13} is a large-scale simulated dataset for evaluating enhancement for additive noise. For the training and validation data, 10 types of noise with SNR = \{15, 10, 5, 0\} are mixed. For the test data, 5 types of unseen noise with SNR = \{17.5, 12.5, 7.5, 2.5\} are mixed. The noise type, interval, and SNR are randomly selected for each clean utterance. We generate the simulated noisy speech as much as the clean Librispeech (i.e., 960, 10, and 10 h for training, validation, and test, respectively). 

CHiME-4 \cite{barker17} provides read speech recorded from noisy environments with a 6-channel tablet microphone. It includes speech with additive noise (4 types) and reverberation. It provides 15, 3, 6, and 5 h of speech for simulted training, real training, validation, and test set, respectively. The acoustic room simulator \cite{chime_simul} is used to generate multi-channel simulated training data which convolve single-channel clean speech with 88 ms impulse response estimated from 65 recordings of tablet microphones, and add 4 types of background noise. During training, the multi-channel data is sampled randomly to make the enhancement model robust to slight changes in source position \cite{du16, dat16}. Among the 6 channels, we report the WER of the 5th channel in the test data.

\subsection{Comparable loss functions}
\label{ssec: Competitors}
As the single channel speech enhancement baseline, we evaluate the Wiener filter method \cite{pascal96}, with smoothing factor $\beta=0.98$. For methods relying on clean speech target, we evaluate the method minimizing the L1 distance between clean and enhanced LMFB feature (DCE), and FSEGAN \cite{donahue17} described in Section II.  

The optimal hyperparameters (i.e., the number of hidden layers and neurons of the models, $(\alpha, \beta)$) were selected based on yielding the minimum WER on validation data, under the DCE loss function. Selected hyperparameters and architecture of $E$ are the same across all of the comparable loss functions.

\begin{figure}
 \centering
  \includegraphics[width=7.5cm]{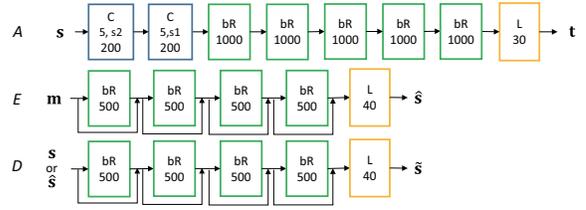}
   \caption{Detailed architectures of acoustic ($A$), enhancement ($E$), and discriminator ($D$). Each box describes the layer type (C: 1D convolutional, bR: bidirectional LSTM-RNN, L: linear) and the kernel size (width, stride, $\#$map) for C, $\#$unit for bR and L.}
 \label{architecture}
\end{figure}

\subsection{Detailed architecture}
\label{sec: architecture}
Fig. \ref{architecture} shows the architectures of each $A$, $E$, $D$ models. The speech feature is employed with the LMFB features. The architecture of $A$ is based on a stack of convolutional and long short-term memory (LSTM) recurrent layers. Each convolutional layer is followed by batch normalization and rectified linear unit nonlinearity. Each recurrent layer is followed by a sequence-wise batch normalization layer \cite{laurent16}. 

Both $E$ and $D$ are multi-layer bidirectional LSTM-RNNs, whose input and output are LMFB features. Moreover, they have a residual connection between the input and output of each layer for better convergence \cite{zhou17}.

\section{Results}
\label{sec: Results}

\begin{figure}
 \centering
  \includegraphics[width=7.5cm]{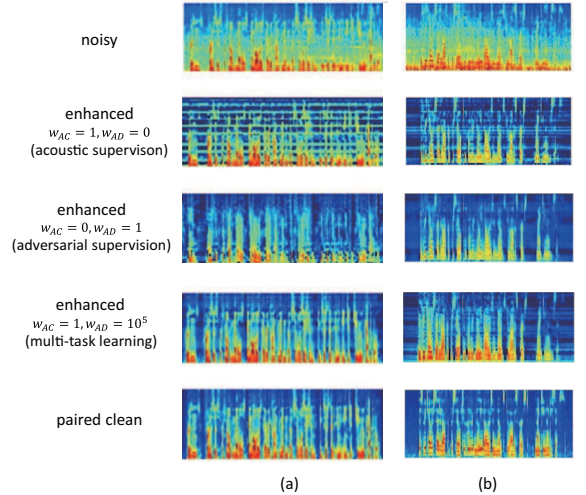}
   \caption{Enhanced test LMFB features obtained using different task combination. (a) Metro noise with SNR=5 in Librispeech+DEMAND. (b) Bus noise with reverberation in CHiME-4}
 \label{generated_samples}
\end{figure}

\subsection{Enhanced feature obtained with different loss functions}
\label{ssec: Generated sample}
Fig.~\ref{generated_samples} shows the LMFB features of noisy, paired clean, and enhanced speech obtained using different loss combinations on the simulated test sets. The enhanced feature obtained using the acoustic supervision ($w_{AC}=1, w_{AD}=0$) contains the characteristic of voice (e.g., harmonics) in the noisy mixture, and artifacts (e.g., the horizontal line for a few frequencies). Compared to acoustic supervision, the enhanced feature obtained using adversarial supervision ($w_{AC}=0, w_{AD}=1$) shows less artifacts, but has less voice characteristic at low frequency. The multi-task learning of AAS ($w_{AC}=1, w_{AD} = 10^5$) maintains voice characteristics in the generated samples while suppressing the artifacts. This tendency is consistently observed in both noisy datasets.

\begin{figure}
 \centering
  \includegraphics[width=7.5cm]{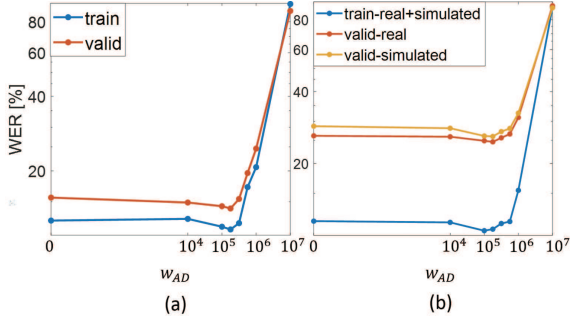}
   \caption{WER with varying loss weight for adversarial supervision (a) on Librispeech + DEMAND, and (b) on CHiME-4}
 \label{WER with DT weight}
\end{figure}

\subsection{WERs and distance between clean and enhanced feature}
\label{ssec: Decoding result}
Fig.~\ref{WER with DT weight} compares the WERs obtained using values of $w_{AD} \in \{0, 10^4, 10^5,10^{5.25}, 10^{5.5}, 10^{5.75}, 10^6, 10^7\}$ given $w_{AC}=1$. 
On both datasets, the lowest WER on the validation data is observed when $w_{AD}$ is between $10^5$ to $10^6$ and starts to increase at some point.

Table \ref{libri+noise} and \ref{chime4-simul} show the WER and DCE (normalized by the number of frames) on the test set of Librispeech + DEMAND, and CHiME-4. The Wiener filtering method shows lower DCE, but higher WER than no enhancement. We conjecture that Wiener filter remove some fraction of noise, however, remaining speech is distorted as well. The adversarial supervision (i.e., $w_{AC}=0, w_{AD}>0$) consistently shows very high WER (i.e., $>$ 90\%), because the enhanced sample tends to have less correlation with noisy speech, as shown in Fig.~\ref{generated_samples}. 

In Librispeech + DEMAND, acoustic supervision (15.6\%) and multi-task learning (14.4\%) achieves a lower WER than minimizing DCE (15.8\%) and FSEGAN (14.9\%). The same tendency is observed in CHiME-4 (i.e. acoustic supervision (27.7\%) and multi-task learning (26.1\%) show lower WER than minimizing DCE (31.1\%) and FSEGAN (29.1\%)).

Because the AM is trained on Librispeech, reducing DCE is directly related to lowering the WER in Librispeech+DEMAND, but does not ensure lowering of the WER in CHiME-4. This explains the slight WER difference between AAS and FSEGAN in Librispeech+DEMAND and the large difference in CHiME-4.

Table \ref{simul_vs_real} shows the WERs on the simulated and real test sets when AAS is trained with different training data. With the simulated dataset as the training data, FSEGAN (29.6\%) does not generalize well compared to AAS (25.2\%) in terms of WER. With the real dataset as the training data, AAS shows
\begin{table}[H]
\centering
\caption{WERs ($\%$) and DCE of different speech enhancement methods on Librispeech + DEMAND test set}
\label{libri+noise}
{\normalsize
\begin{tabular}{c|c|c}
\hline
Method               & WER (\%) & DCE \\ \hline
No enhancement       & 17.3          & 0.828 \\
Wiener filter & 19.5          & 0.722 \\ \hline
Minimizing DCE    & 15.8          & \textbf{0.269} \\
FSEGAN               & 14.9          & 0.291 \\ \hline
AAS \begin{scriptsize}($w_{AC}=1,w_{AD}=0$)\end{scriptsize}                  & 15.6          & 0.330 \\
AAS \begin{scriptsize}($w_{AC}=1,w_{AD}=10^5$)\end{scriptsize}                   & \textbf{14.4}          & 0.303 \\ \hline
Clean speech         & 5.7           & 0.0   \\ \hline
\end{tabular}
}
\end{table}
\begin{table}[H]
\centering
\caption{WERs ($\%$) and DCE of different speech enhancement methods on CHiME4-simulated test set}
\label{chime4-simul}
\normalsize
\begin{tabular}{c|c|c}
\hline
Method               & WER (\%) & DCE \\ \hline
No enhancement       & 38.4          & 0.958 \\
Wiener filter & 41.0          & 0.775 \\ \hline
Minimizing DCE    & 31.1          & \textbf{0.392} \\
FSEGAN               & 29.1          & 0.421 \\ \hline
AAS \begin{scriptsize}($w_{AC}=1,w_{AD}=0$)\end{scriptsize}                  & 27.7          & 0.476 \\
AAS \begin{scriptsize}($w_{AC}=1,w_{AD}=10^5$)\end{scriptsize}                   & \textbf{26.1}          & 0.462 \\ \hline
Clean speech         & 9.3           & 0.0   \\ \hline
\end{tabular}
\end{table}
% \usepackage{multirow}
\begin{table}[H]
\centering
\caption{WERs ($\%$) of obtained using different training data of CHiME-4}
\label{simul_vs_real}
\normalsize
\begin{tabular}{c|c|c|c}
\hline
\multirow{2}{*}{Method}                                                & \multirow{2}{*}{Training Data} & \multicolumn{2}{c}{Test WER (\%)} \\ \cline{3-4} 
                                                                       &                                & simulated          & real          \\ \hline
\multirow{3}{*}{\begin{tabular}[c]{@{}c@{}}AAS\\ \begin{scriptsize}($w_{AC}=1, w_{AD}=10^5$)\end{scriptsize}\end{tabular}} & simulated                      &                 26.1 &   25.2         \\
                                                                       & real                           &   37.3              &  35.2          \\
                                                                       & simulated + real                &   25.9              & 24.7           \\ \hline
FSEGAN                                                                 & simulated                      & 29.1                & 29.6           \\ \hline
\end{tabular}
\end{table}
\noindent
severe overfitting since the size of training data is small. When AAS is trained with simulated and real datasets, it achieves the best result (24.7\%) on the real test set.
\section{Conclusion}
\label{sec:conclusion}
Speech enhancement models have several limitations when using clean speech from the simulated database as the target. To avoid relying on clean speech target, we propose training speech enhancement model with the multi-task learning of acoustic and adversarial supervision (AAS). Each supervision maximizes the likelihood of transcription on the pre-trained acoustic model and ensures general characteristics of clean speech in the enhanced output, which improves generalization on unseen noisy speech. The proposed method was tested on two datasets: Librispeech + DEMAND and CHiME-4. By visualizing the enhanced feature, we demonstrated the role of each supervision. AAS showed a lower word error rate compared to speech enhancement methods using a clean target. The proposed AAS can be combined with any acoustic model of a given clean speech and noisy speech with transcription.

%\ifCLASSOPTIONcaptionsoff
%  \newpage
%\fi

\bibliographystyle{IEEEtran}
\newpage

\bibliography{references}
\label{sec:ref}

\end{document}